\documentclass[sigconf]{acmart}
\usepackage{colortbl} 
\usepackage{booktabs} 
\usepackage{textcomp}
\usepackage{stfloats}
\usepackage{url}
\usepackage{verbatim}
\usepackage{graphicx}
\usepackage{multirow}
\usepackage{soul}
\usepackage{tabularray}
\usepackage[normalem]{ulem}
\useunder{\uline}{\ul}{}
\usepackage{tabularx}
\usepackage{adjustbox}
\usepackage{enumitem}
\usepackage{hyperref}

\setcopyright{none}
\renewcommand\footnotetextcopyrightpermission[1]{}
\AtBeginDocument{%
  }

\begin{document}

\title{Residual Prior-driven Frequency-aware Network for Image Fusion}

\author{Zheng Guan}
\email{ guanzheng@ynu.edu.cn}
\affiliation{%
  \institution{The School of Information Science and Engineering, Yunnan University}
  \city{Kunming}
  \state{Yunnan}
  \country{China}
}

\author{Xue Wang}
\authornote{Corresponding Author}
\email{wangxue2022@mail.ynu.edu.cn}
\affiliation{%
  \institution{The School of Information Science and Engineering, Yunnan University}
  \city{Kunming}
  \state{Yunnan}
  \country{China}
}
\author{Wenhua  Qian}
\email{whqian@ynu.edu.cn}
\affiliation{%
  \institution{The School of Information Science and Engineering, Yunnan University}
  \city{Kunming}
  \state{Yunnan}
  \country{China}
}
\author{Peng Liu}
\email{20250028@ynu.edu.cn}
\affiliation{%
  \institution{The School of Information Science and Engineering, Yunnan University}
  \city{Kunming}
  \state{Yunnan}
  \country{China}
}
\author{Runzhuo Ma}
\email{runzhuo.ma@connect.polyu.hk}
\affiliation{%
  \institution{The Department of Electrical and Electronic Engineering, Hong Kong Polytechnic University,}
  \city{Hong Kong}
  \country{China}
}


\renewcommand{\shortauthors}{Guan et al.}

\begin{abstract}

Image fusion aims to integrate complementary information across modalities to generate high-quality fused images, thereby enhancing the performance of high-level vision tasks. While global spatial modeling mechanisms show promising results, constructing long-range feature dependencies in the spatial domain incurs substantial computational costs. Additionally, the absence of ground-truth exacerbates the difficulty of capturing complementary features effectively. To tackle these challenges, we propose a Residual Prior-driven Frequency-aware Network, termed as RPFNet. Specifically, RPFNet employs a dual-branch feature extraction framework: the Residual Prior Module (RPM) extracts modality-specific difference information from residual maps, thereby providing complementary priors for fusion; the Frequency Domain Fusion Module (FDFM) achieves efficient global feature modeling and integration through frequency-domain convolution. Additionally, the Cross Promotion Module (CPM) enhances the synergistic perception of local details and global structures through bidirectional feature interaction. During training, we incorporate an auxiliary decoder and saliency structure loss to strengthen the model’s sensitivity to modality-specific differences. Furthermore, a combination of adaptive weight-based frequency contrastive loss and SSIM loss effectively constrains the solution space, facilitating the joint capture of local details and global features while ensuring the retention of complementary information. Extensive experiments validate the fusion performance of RPFNet, which effectively integrates discriminative features, enhances texture details and salient objects, and can effectively facilitate the deployment of the high-level vision task.
\textit{ The source code
can be available at \href{https://github.com/wang-x-1997/RPFNet}{https://github.com/wang-x-1997/RPFNet}.}

\textit{\textbf{This paper has been accepted by ACM MM 2025}}
\end{abstract}




\begin{teaserfigure}
  \includegraphics[width=\textwidth]{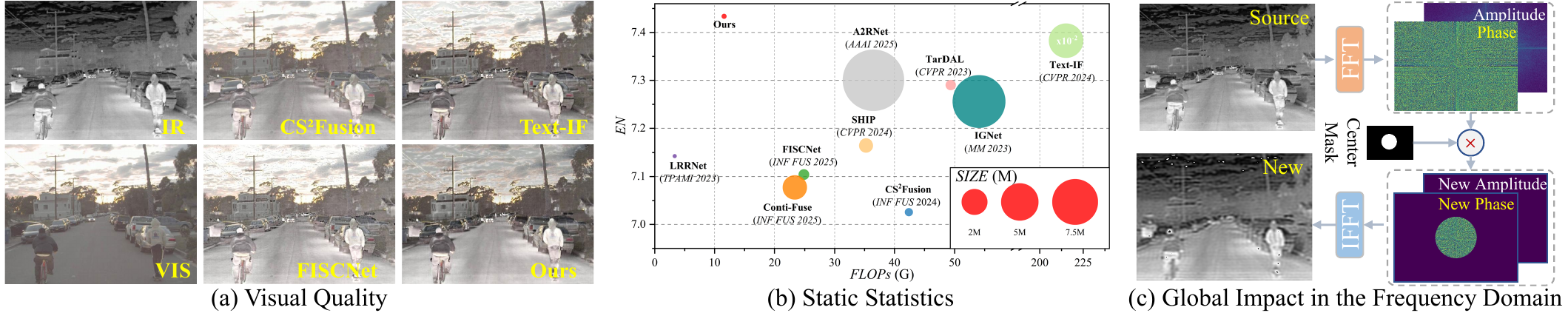}
  \caption{Comparison of SOTA methods and our method on the FLIR dataset. (a) Three representative methods are considered: CS$^2$Fusion \cite{4} (CNN with contrastive learning), FISCNet \cite{5} (CNN with frequency transformation), and Text-IF \cite{7} (Transformer with semantic text). In comparison, our method demonstrates more competitive visual results. (b) Quantitative comparison of entropy (EN) and computational efficiency (SIZE and FLOPs), showing our method achieves high-quality fusion with lower computational complexity. (c) Taking infrared images as an example, local modifications to the phase component of Fast Fourier transform (FFT) result in global changes in the spatial domain after inverse FFT (IFFT), validating the effectiveness of frequency domain processing for global feature modeling.}
  \label{fig1}
\end{teaserfigure}


\maketitle

\section{Introduction}
\label{sec1}
Image fusion serves as an enhancement technique aimed at integrating complementary scene information captured by multiple sensors, generating information-rich fused images that overcome the limitations of single sensors and facilitate the deployment of high-level vision tasks \cite{2,3,53,35,56}. Infrared and visible image fusion (IVIF) represents a critical research domain that combines the thermal information from infrared (IR) images with the rich textural details of visible (VIS) images, producing comprehensive scene representations that mitigate the illumination sensitivity of visible images and the noise susceptibility and low resolution of infrared images. This comprehensive information integration provides solutions for a wide range of scenarios and requirements \cite{4,5,6,7,8,9,10,11,12}.

With the advancement of deep learning (DL) techniques, IVIF has gained more effective guidance, particularly when addressing challenges posed by various environments and conditions. A typical pipeline that utilizes the nonlinear capabilities of CNNs to learn fusion maps has shown promising results due to its strong generalization ability. However, CNNs have limitations in their ability to model long-range dependencies, as they only capture interactions within their receptive fields, which presents challenges when balancing texture and saliency objectives. To enhance fusion performance, Transformer-based architectures were introduced, utilizing self-attention mechanisms to capture global contextual interactions and model long-range dependencies. Nevertheless, these methods incur high computational complexity and parameter counts, with self-attention computation costs growing quadratically when processing complex images, limiting their application in resource-constrained environments. Figure \ref{fig1} (a) and (b) provide intuitive examples. Furthermore, the lack of direct ground truth complicates the learning of fusion maps, as existing methods typically rely on proxy losses. In this context, effectively capturing and integrating complementary information from different modalities remains challenging.

To tackle these challenges, we propose a Residual Prior-driven Frequency-aware Network (RPFNet) based on the following key observations:  1) Residual maps between modalities contain critical textural priors \cite{50,51}; 2) Frequency domain processing efficiently captures global information compared to spatial domain long-range modeling (Figure~\ref{fig1}(c)); 3) Coupled convolution across spatial-frequency domains facilitates collaborative local-global feature perception \cite{5,6,30}. Specifically, RPFNet features a dual-branch architecture: a Residual Prior Module (RPM) capturing fine-grained multi-scale features from residual maps, and a Frequency Domain Fusion Module (FDFM) efficiently integrating global information via Fourier transformation. Moreover, a Cross-Promotion Module (CPM) enhances bidirectional interaction between RPM and FDFM. RPM provides residual priors to guide frequency fusion, while FDFM feedback refines residual feature perception. During training, an auxiliary decoder with a designed saliency structural loss $\mathcal{L}_{SS}$ strengthens modality-specific residual difference perception. An adaptive weight-based frequency contrastive fidelity term $\mathcal{L}_{c}$ shrinks solution space through frequency-domain contrastive learning, promoting global awareness. Coupled with local SSIM constraints \cite{39}, our model effectively captures complementary local-global features. Extensive experiments validate the method’s fusion performance and utility in high-level vision tasks.
In summary, our contributions include:
\begin{itemize}
\item We propose a novel fusion framework that effectively combines the advantages of residual priors and frequency domain processing, achieving computationally efficient global feature modeling while ensuring effective preservation of local features.
\item Efficient modules, such as CPM designed to enhance the perception of complementary features, improve local details, and preserve feature structures, are proposed to boost the performance of RPFNet.
\item We introduce a saliency structure loss and an adaptive weight-based frequency contrastive fidelity term. The former enhances the model's perception of modality differences, while the latter collaborates with SSIM to facilitate the capture of local-global features, ensuring the preservation of complementary information.
\end{itemize}


\section{ Related work}
\textbf{DL-based Image Fusion.} Early DL-based methods primarily focused on constructing CNN architectures to integrate complementary cross-modal contextual features \cite{2,3}. These methods can be categorized into three main paradigms based on their training strategies: end-to-end models \cite{4,5,6,8,10}, autoencoder (AE)-based methods \cite{1,9,11,12,22}, and generative models \cite{13,14,17}. End-to-end models typically rely on the effective combination of network architectures and loss functions to drive the model to adaptively preserve complementary information faithful to source images. AE-based methods leverage the comprehensive feature extraction capabilities of autoencoders coupled with carefully designed fusion strategies to achieve effective integration of cross-modal features. As a classical generative framework, GAN establish adversarial games between inputs and outputs to ensure effective preservation of source features while maintaining perceptual satisfaction. While these methods have shown promising progress, their CNN-based architectures limit them to modeling interactions only within their receptive fields, lacking effective long-range dependency modeling \cite{18,19}. The self-attention mechanism in the Transformer provides an effective solution to this limitation \cite{18,19,7,21,23}. Tang et al. \cite{18} leveraged Transformers to model both intra-domain and inter-domain long-range dependencies, facilitating cross-modal feature interactions. Tang et al. \cite{19} proposed a dynamic Transformer to construct long-term relationships, promoting thorough fusion of local-global complementary features. Wang et al. \cite{21} employed self-supervised masked image modeling to enhance local feature extraction and global consistency. Yi et al. \cite{7} constructed a text-driven Transformer to help fusion networks address multiple degradation issues. Despite their effectiveness, these methods often incur high computational costs \cite{4,6,8,9}. Unlike previous approaches, our proposed RPPNet captures modal difference information through residual priors and employs frequency domain transformations for global feature modeling, thereby achieving high-quality feature fusion while maintaining computational efficiency.

\begin{figure*}[]
    \centering
    \includegraphics[scale=0.49]{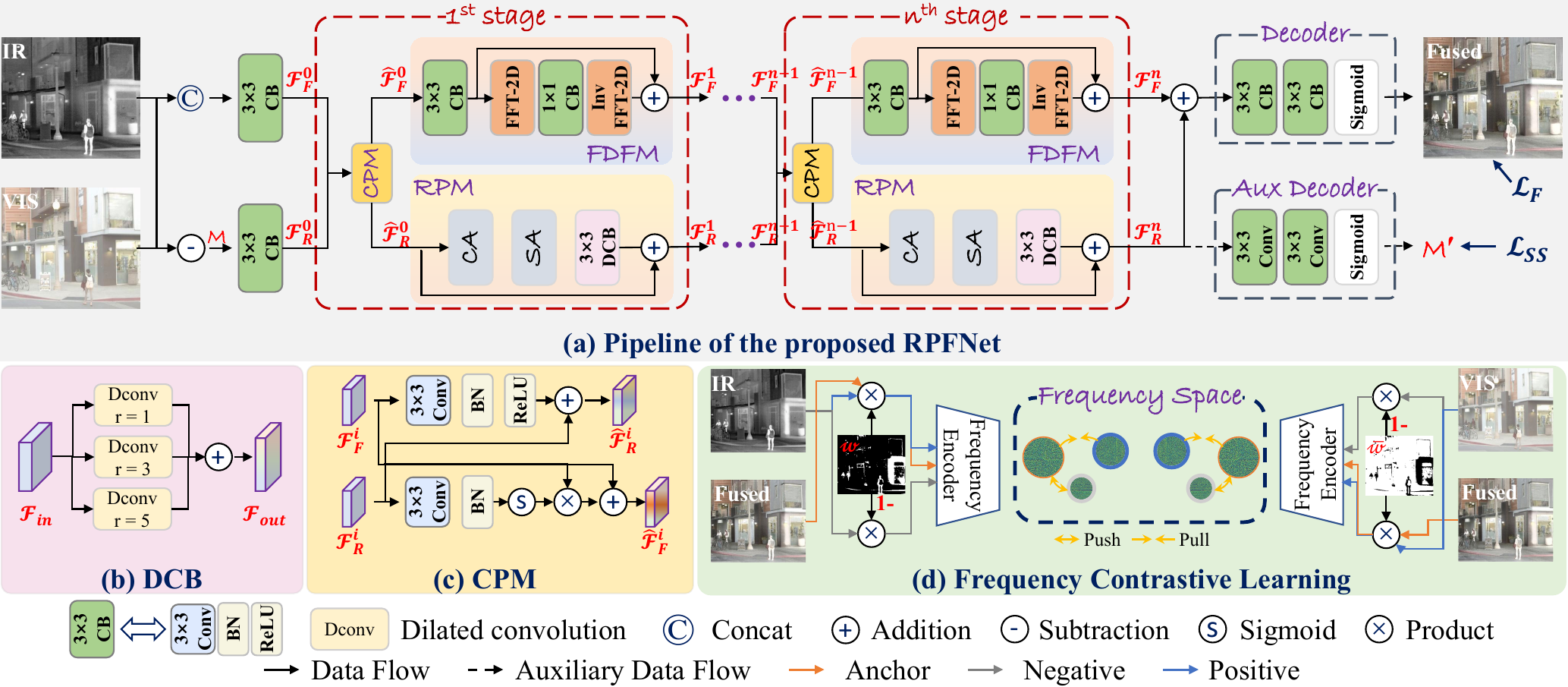}
    \caption{
    Overview of the Proposed Residual Prior-driven Frequency-aware Network.
    }
    \label{fig2}
\end{figure*}

\textbf{Frequency Domain Learning.} Recent advancements, such as Fast Fourier Convolution \cite{24}, leverage frequency domain techniques for efficient global feature map, inspiring various frequency-based methods \cite{24,29,55,57,58}. For instance, Kong et al.~\cite{27} designed frequency-domain attention to simplify models; Chen et al.~\cite{28} introduced frequency-adaptive convolution for enhanced dilated operations. Furthermore, frequency domain robustness has advanced image restoration tasks like denoising \cite{32,33}, deraining \cite{30,31}, and deblurring \cite{27}. In IVIF, Zheng et al.~\cite{5} mined complementary features using spatial-frequency synergies, and \cite{6,34} explored higher-order interactions to integrate fine-grained spatial details with global statistics.

\textbf{Contrastive Learning for Image Fusion.} Contrastive learning effectively captures implicit feature relationships without ground truth by constructing positive-negative sample pairs \cite{4,35,36,37,52}. Wang et al.~\cite{4} leveraged feature similarity-divergence to extract complementary modal features, while Liu et al.~\cite{36} employed annotations balancing salient and textural features. Zhu et al.~\cite{37} proposed patch-based positive-negative sampling for feature preservation, and Wang et al.~\cite{35} utilized unpaired high-quality images for degradation-resistant fusion.

\section{Methodology}
\subsection{Overview}
RPFNet adopts a multi-stage architecture with iterative optimization, as illustrated in Figure \ref{fig2}(a). Each iteration stage consists of three collaboratively working core modules: FDFM, RPM, and CPM. Specifically, FDFM employs the FFT to map features into the frequency domain, enabling efficient global feature modeling at a reduced computational cost. This addresses the limitations of CNNs in capturing long-range dependencies and avoids the quadratic complexity of Transformer-based self-attention mechanisms; RPM extracts multi-scale structural information from the residual map $\mathcal{M} := \text{IR}-\text{VIS}_Y$, where $\text{VIS}_Y$ is the luminance channel of the visible image. By focusing on modal differences, RPM provides a prior that guides the fusion process to preserve critical thermal and textural details; CPM facilitates bidirectional feature interaction between FDFM and RPM. It enhances the residual features using contextual information from the fused features and, conversely, refines the fused features by incorporating modality-specific priors from the residual branch. This cross-modal propagation ensures that the model effectively integrates complementary information across modalities. After $N$ iterations, the model generates a fused image. Additionally, an auxiliary decoder is introduced during training, utilizing a saliency structure loss $\mathcal{L}_{SS}$ to enhance the model’s ability to capture modality-specific differences in the residual map $\mathcal{M}$.  The advantages of this architecture are threefold: 1) The introduction of residual priors enables the model to learn effective fusion strategies from modal differences, effectively alleviating the lack of supervision signals; 2) Feature processing based on frequency-domain transformation not only achieves natural long-range dependency modeling but also significantly reduces computational complexity; 3) The iterative optimization mechanism realizes progressive improvement of fusion performance through a closed-loop design of \textit{"extraction-interaction-update"}.

\subsection{ Network Architecture}
Given a pair of input images $\{ \text{IR}\in \mathbb{R}^{H \times W \times1} , \text{VIS} \in \mathbb{R}^{H \times W \times3}  \}$, the processing pipeline begins by converting $\text{VIS}$ to the $YCbCr$ color space and extracting its luminance channel ($Y$) as $\text{VIS}_{Y}$. The inputs $\text{IR}$ and $\text{VIS}_{Y}$   are then processed in two parallel pathways: 1) Concatenated and fed into a Convolutional Block (CB) \cite{4} to extract coarse fused features $\mathcal{F}_{F}^{0}$; 2) Used to compute a residual map $\mathcal{M}$, from which salient structural information $\mathcal{F}_{R}^{0}$ is extracted via another CB. These initial features, $\mathcal{F}_{F}^{0}$ and $\mathcal{F}_{R}^{0}$, are iteratively refined over $N$ stages. For clarity, we describe the operations within the first iteration ($N=1$), with subsequent iterations following the same structure.

\textbf{Cross-Promotion Module (CPM):} The architecture of CPM, as illustrated in Figure \ref{fig2} (c), is designed to facilitate bidirectional enhancement between fusion and residual features. It operates as follows: 1) The rough fusion feature $\mathcal{F}_{F}^{0}$ is processed by a 3$\times$3 convolution, BN, and ReLU, and the output is added to the residual features $\mathcal{F}_{R}^{0}$ via a skip connection. This enriches the residual features with contextual information from the fused branch: 
\begin{equation}
\widehat{\mathcal{F}}_R^0=\operatorname{ReLU}\left(\operatorname{B N}\left(\operatorname{Conv}_{3 \times 3}\left(\mathcal{F}_F^0\right)\right)\right)+\mathcal{F}_R^0.
\end{equation}
2) The residual features $\mathcal{F}_{R}^{0}$ are transformed into an attention map using a 3$\times$3 convolution, BN, and Sigmoid activation. This map is then applied to the fused features $\mathcal{F}_F^0$ via element-wise multiplication, emphasizing regions with significant modality differences:
\begin{equation}
\widehat{\mathcal{F}}_F^0=\delta\left(\operatorname{B N}\left(\operatorname{Conv}_{3 \times 3}\left(\mathcal{F}_R^0\right)\right)\right) \cdot \mathcal{F}_F^0+\mathcal{F}_F^0,
\end{equation}
where $\delta$ denotes the Sigmoid function.

\textbf{Residual Prior Module (RPM):} The refined residual features $\widehat{\mathcal{F}}_R^0$ are fed into RPM, which comprises Channel Attention (CA) \cite{38}, Spatial Attention (SA) \cite{38}, and Dilated Convolutional Block (DCB). RPM aims to extract multi-scale contextual information from $\widehat{\mathcal{F}}_R^0$ using dilated convolutions with varying receptive fields, enhancing its role as a prior for feature fusion. Due to modality differences, residual priors often contain noise, which can be amplified during encoding. To address this, CA and SA refine $\widehat{\mathcal{F}}_R^0$ with fine-grained attention, followed by dilated convolutions with dilation rates $r \in \{ 1, 3, 5 \}$ to capture multi-scale contexts. The updated residual features are computed as:
\begin{equation}
\mathcal{F}_R^1=\sum_{r \in\{1,3,5\}} {DCov}_r\left({SA}\left({CA}\left(\hat{\mathcal{F}}_R^0\right)\right)\right),
\end{equation}
where ${DCov}_r$ denotes dilated convolution with dilation rate $r$.

\begin{figure}[]
    \centering
    \includegraphics[scale=0.29]{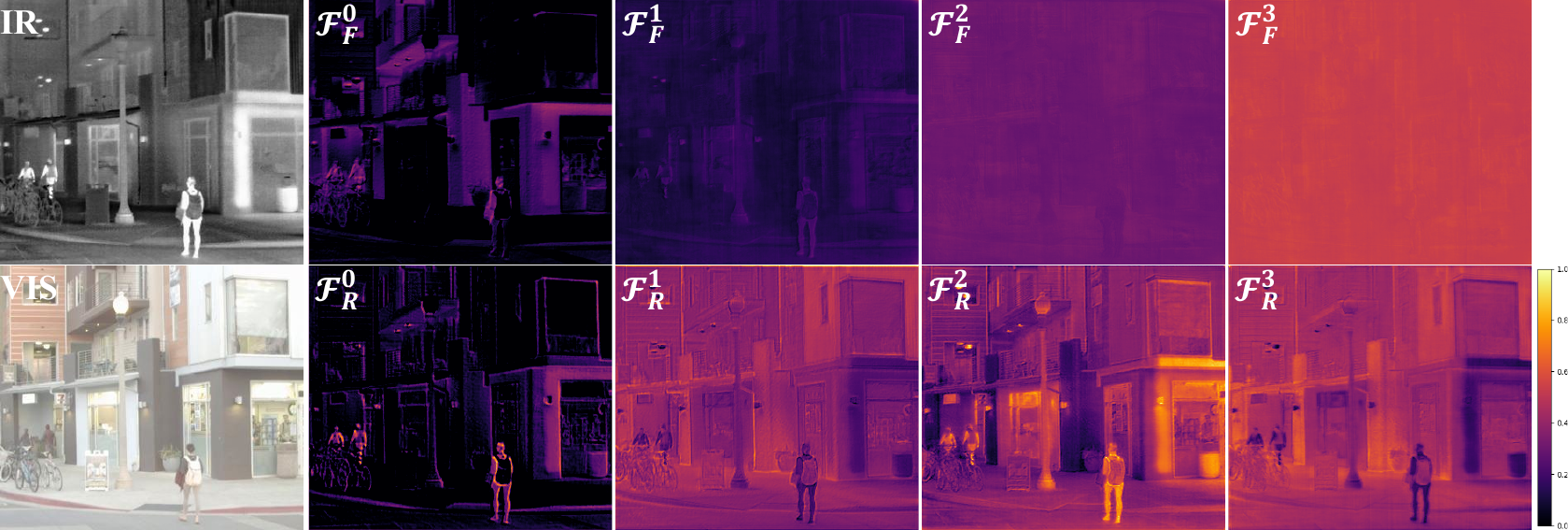}
    \caption{
    A visualization example of feature maps in RPFNet demonstrates the progressive optimization process from input to output. The FDFM effectively integrates global contextual information through frequency-domain transformations, while the RPM focuses on extracting complementary features from the $\mathcal{M}$, emphasizing modal-specific information. All maps have been normalized, and it is recommended to use appropriate color maps to achieve optimal readability.
    }
    \label{fig3}
\end{figure}

\textbf{Frequency-Domain Feature Module (FDFM):} FDFM processes $\widehat{\mathcal{F}}_F^0$ in the frequency domain for efficient global modeling. The workflow consists of three steps: 1) a $3 \times 3$ CB refines to enhance fusion information. 2) The refined features undergo FFT to transform them into the frequency domain, followed by a $1 \times 1$ CB for frequency-domain convolution, equivalent to global spatial convolution at a lower computational cost. 3) IFFT converts the features back to the spatial domain, yielding globally modeled fused features. Additionally, a parallel spatial branch with a residual connection preserves local details. The FDFM output is expressed as:
\begin{equation}
\mathcal{F}_F^1={CB}_{3 \times 3}(\hat{\mathcal{F}}_F^0)+\mathcal{FFT}^{-1}({CB}_{1 \times 1}(\mathcal{FFT}({CB}_{3 \times 3}(\hat{\mathcal{F}}_F^0)))),
\end{equation}
where $\mathcal{FFT}$ denotes the FFT and $\mathcal{FFT}^{-1}$ denotes the IFFT. Figure  \ref{fig3} presents the visualization of feature maps at different stages, intuitively demonstrating the design rationality of the proposed method.

\textbf{Iterative Refinement and Final Fusion:} The outputs $\mathcal{F}_F^1$ and $\mathcal{F}_R^1$ serve as inputs for the next iteration. After $N$ iterations, the refined features
$\mathcal{F}_F^n$ and $\mathcal{F}_R^n$ are summed and passed through a reconstruction head, consisting of two CBs and a Sigmoid activation, to generate the final fused image. The chrominance of the fused image is restored using the YCbCr color space. During training, an auxiliary decoder applies the Saliency Structure Loss $\mathcal{L}_{SS}$ to better capture modal differences by constraining $\mathcal{M}$ (or its derived representations) to align with the expected properties of residual priors. $\mathcal{L}_{SS}$ directly leverages inter-modal differences rather than relying solely on output-to-input similarity metrics. It consists of two components:
\begin{equation}
\mathcal{L}_{SS}=\mathcal{L}_{{grad }}+\mathcal{L}_{{reg }}.
\end{equation}
Each term imposes specific constraints on $\mathcal{M}$ to maintain consistency with the desired attributes of residual priors. $\mathcal{L}_{grad}$
 preserves thermal edge features by aligning the gradients of $\mathcal{M}^{\prime}$ and IR, while emphasizing regions with high texture complexity to ensure detail preservation from VIS:
 \begin{equation}
\mathcal{L}_{{grad }}=\left\|\nabla\left(\mathcal{M}^{\prime}\right)-\nabla(I R)\right\|_2- \lambda_{1}\cdot \|\mathcal{M}^{\prime} \cdot \mathfrak{m}(V I S_{Y})\|_{2},
\label{eq6}
\end{equation}
where $\lambda_1$ is a trade-off parameter, $\nabla$ denotes the Sobel operator, $\|\cdot\|_2$ represents the ${l}_2$ norm. $\mathfrak{m}=\mathbb{I}\{s(V I S_{Y})>\mu(s(V I S_{Y}))\}$ represents a texture mask that emphasizes gradient information, where $\mathbb{I}$ is the indicator function and $\mu$ denotes the mean operation. $\mathcal{L}_{reg }$ focuses on thermally salient information in IR images, employing IR's statistical properties to encourage response in these regions:
\begin{equation}
\mathcal{L}_{{reg }}=-\|\mathcal{M}^{\prime} \cdot T(I R)\|_{2},
\end{equation}
where $T=\mathbb{I}\{I R>(\mu(I R)+\sigma(I R))\}$, with $\sigma$ denotes the standard deviation operation.

\subsection{Object Function}
To drive the model to better preserve complementary information from source images, the fusion loss $\mathcal{L}_{F}$ comprises two key components: frequency domain contrastive loss and structural loss, defined as:
\begin{equation}
\mathcal{L}_{F}=\mathcal{L}_{{c }}+ \lambda_{2}\cdot\mathcal{L}_{{s}},
\label{eq8}
\end{equation}
where $\lambda_2$ is a trade-off parameter. In existing DL-driven image fusion methods, loss functions such as $\ell_1/\ell_2$ distance and weighted fidelity loss have been widely applied. However, these loss functions typically focus on similarities of local pixels or small regions, making it difficult to effectively constrain global structural features. Although some methods have improved global feature capture by constructing contrastive perceptual regularization terms, they mostly focus on spatial domain feature information while neglecting the rich global information contained in the frequency domain.
Specifically, in the IVIF task, low-frequency components in infrared images (such as overall distribution of thermal radiation) and high-frequency components in visible images (such as texture and edge details) together constitute complementary features. Therefore, constructing contrastive loss in the frequency domain can more naturally capture the distributional characteristics of these high-frequency and low-frequency features.

In particular, we design adaptive weights $w$ based on the inherent characteristics of IR and inter-modal difference information, and reference \cite{36} to mark the foreground saliency and background texture features of source images to construct positive and negative sample pairs. The calculation of the weight $w$ proceeds as follows:
\begin{equation}
\begin{gathered}
\mathcal{S}_{IR}=\delta\left(\frac{IR-\mu(IR)}{\sigma(IR)}\right), 
\mathcal{S}_{\mathcal{M}}=\delta\left(\frac{\mathcal{M}-\mu(\mathcal{M})}{\sigma(\mathcal{M})}\right), \\
\mathcal{S}_{en}=\left(\mathcal{S}_{IR} \cdot\left(1-\mathcal{S}_{\mathcal{M}}\right)+\mathcal{S}_{\mathcal{M}} \cdot\left(1-\mathcal{S}_{IR}\right)\right) / 2 ,\\
w=\mathbb{I}\left\{\mathcal{S}_{en}>\left(\mu\left(\mathcal{S}_{en}\right)+\sigma\left(\mathcal{S}_{en}\right)\right)\right\}.
\end{gathered}
\end{equation}
Considering the frequency domain features' capacity for global feature description (see Figure  \ref{fig1}(c)), we employ FFT for feature extraction and measure the $\ell_1$ distance between samples in the frequency domain to achieve global fusion control, rather than being limited to local receptive fields and spatial domain features as in VGG-based methods. Figure \ref{fig2}(d) illustrates the frequency contrastive learning process, which can be formulated as:
\begin{equation}
\mathcal{L}_{c}=\frac{\mathcal{D}_{{pos }}}{\mathcal{D}_{{neg }}+\epsilon},
\end{equation}
\begin{equation}
\begin{aligned}
\mathcal{D}_{{pos }} &=  \|\mathcal{FFT}(F \cdot w) - \mathcal{FFT}(I R \cdot w)\|_{1} \\&+\|\mathcal{FFT}(F \cdot \bar{w})-\mathcal{FFT}(V I S_{Y} \cdot \bar{w})\|_{1}, \\ 
\mathcal{D}_{ {neg }} &=\sum_{(a, b) \in \mathcal{N}} \|\mathcal{FFT}(a), \mathcal{FFT}(b)\|_{1},
\end{aligned}
\end{equation}
Here, $\epsilon$ is set to 1e$^{-9}$ to prevent gradient collapse, $\|\cdot\|_{1}$ is the $\ell_1$ norm, $F$ denotes the $Y$ component of the fused image, ${\bar{w}} = 1 - {w}$ is the complement of IR, and $\mathcal{N}$ denotes the set of all negative sample pairs:
\begin{equation}
\begin{aligned}
\mathcal{N}=&\{(F \cdot w, I R \cdot {\bar{w}}),(F \cdot w, V I S_{Y} \cdot w),\\&(F \cdot {\bar{w}}, I R \cdot w),(F \cdot \bar{w}, V I S_{Y} \cdot \bar{w})\}.
\end{aligned}
\end{equation}

Furthermore, to drive the model to better perceive local details within global features, we introduce a structural regularization term using SSIM function, which aims to maintain the structural integrity of the fused image through its local operation of the sliding window:
\begin{equation}
\mathcal{L}_{s}=\mathcal{SSIM}(F,IR) +\mathcal{SSIM}(F,V I S_{Y}) ,
\end{equation}
where $\mathcal{SSIM}(\cdot)$ denotes the SSIM operation \cite{39}.
Finally, the total objective function for the entire model is formulated as:
\begin{equation}
\mathcal{L}_{total} = \mathcal{L}_{SS} + \mathcal{L}_{F}.
\end{equation}

\section{Experiments and Results}

\subsection{Experimental Setting}
\subsubsection{Datasets and Benchmark.} Four publicly available datasets were utilized for training, validation, and testing in our experiments: FLIR \cite{40}, TNO \cite{41}, M$^3$FD \cite{13}, and MSRS \cite{42}. Specifically, we divided the FLIR dataset, which contains complex urban street environments, into 180 image pairs for train and \textbf{40} pairs for validation. To enhance the training of our RPFNet, we augmented the train dataset through random cropping, expanding it to 3,383 image pairs with size of 128$\times$128. Additionally, we randomly selected \textbf{40} pairs from the military scenario dataset TNO, and incorporated the multi-scene MSRS dataset (containing \textbf{361} pairs) and the \textit{'independent scene fusion'} subset from the M$^3$FD dataset (containing \textbf{300} pairs) into our test dataset to validate the proposed method.

To evaluate the effectiveness and generalizability of our method, we compared it with \textbf{9} SOTA methods: LRRNet \cite{9}, CS$^2$Fusion \cite{4}, Text-IF \cite{7}, FISCNet \cite{5}, TarDAL \cite{13}, IGNet \cite{8}, SHIP \cite{6}, Conti-Fuse \cite{10}, and A$^2$RNet \cite{14}. We employed \textbf{6} metrics: three non-reference metrics (EN, SF, SD) and three reference metrics (CC, SCD, and VIF). For all these metrics, higher values indicate better fusion performance \cite{2,3}.

Furthermore, we partitioned the M$^3$FD dataset—which provides \textit{'for fusion, detection, and fused-based detection'}—into train, validation, and test  datasets following a ratio of 0.6-0.2-0.2. YOLOv5 \cite{46} was employed as the detector to validate our method's performance in advancing high-level vision tasks. For fair comparison, we retained detection models for fusion results from all 9 SOTA methods, with mAP@.5 measured for quantitative comparison.

Additionally, we extended our model to the medical image fusion (MIF) task to verify its versatility. Specifically, we selected 217 image pairs from the Harvard Medical School website \cite{47} as our train dataset, applying the same training standards and evaluation metrics as in the IVIF task. Our test dataset consisted of 10 MRI-PET and 20 MRI-SPECT image pairs. For fair comparison, we benchmarked RPFNet against three SOTA MIF methods: MATR \cite{43}, MMIF-INet \cite{44}, and MMNet \cite{45}.
\subsubsection{Implementation Details.}
Our RPFNet was deployed on an NVIDIA 2080Ti GPU with 11GB memory, paired with a 3.0 GHz Intel i7-9700 CPU. During training, we used two Adam optimizers with identical parameter settings to train the RPFNet and auxiliary decoder separately. The configuration included an initial learning rate of 10$^{-3}$ that decreased by 50\% every 20 epochs, with a batch size of 16. Based on experience, we set the parameter $\lambda_1$ in Eq.(\ref{eq6}) to 0.3, while parameter $\lambda_2$ in Eq.(\ref{eq8}) and the number of stages $n$  were set to 5 and 3 respectively after validation experiments.

\subsection{Comprehensive performance evaluation}
This section presents a comprehensive evaluation of our proposed method, including parameter analysis, ablation studies, and experimental results on the test dataset. These analyses provide intuitive insights into the model’s effectiveness and robustness across diverse experimental settings.
\subsubsection{Parameters analysis}
In our model, two key parameters, the number of stages $N$ and the trade-off parameter $\lambda_2$ in Eq.(\ref{eq8}), play a critical role in guiding fusion performance. To determine optimal values that balance efficacy and efficiency, we conducted the parameter analysis on the validation dataset.

\textbf{Number of Stages $N$:} We investigated the impact of varying the number of stages, testing configurations with 1, 3, 5, 7, and 9 stages. The results, reported in Table \ref{tab1}, reveal that under the residual prior and frequency-domain global modeling framework, the model effectively captures complementary features from source images, achieving the best overall performance at $N=3$. However, further increasing $n$ leads to a declining trend in metrics (except SD), potentially due to challenges associated with gradient propagation. Thus, considering both performance and computational efficiency, we set $N=3$.

\begin{table}[]
\caption{Quantitative results with the different numbers of stages $n$ on the validation dataset. RoR \cite{4,35} provides an intuitive and comprehensive comparison by ranking the average performance rankings of different methods across various metrics. The best values are highlighted in bold.}
\centering
\renewcommand\arraystretch{1.2}
\setlength\tabcolsep{7pt} 
\resizebox{\linewidth}{!}{
\begin{tabular}{cccccccc}
\hline
\rowcolor{gray!15} 
\textbf{$N$} & EN & SF & SD & CC & SCD & VIF & RoR \\ \hline
1 & \textbf{7.459} & \textbf{6.581} & 10.68 & 0.640 & 1.728 & 0.668 & 2 \\
\rowcolor{gray!5} 
3 & 7.434 & 6.235 & 10.69 & 0.642 & \textbf{1.736} & \textbf{0.675} & 1 \\
5 & 7.415 & 6.150 & 10.76 & 0.642 & 1.734 & 0.660 & 3 \\
\rowcolor{gray!5} 
7 & 7.379 & 6.121 & 10.45 & \textbf{0.645} & 1.727 & 0.662 & 4 \\
9 & 7.369 & 5.958 & \textbf{10.77} & 0.634 & 1.685 & 0.636 & 5 \\ \hline
\end{tabular}
}
\label{tab1}
\end{table}
\begin{table}[]
\caption{Quantitative results with the different trade-off parameter $\lambda_{2}$ on the validation dataset. The best values are highlighted in \textbf{bold}.}
\centering
\renewcommand\arraystretch{1.2}
\setlength\tabcolsep{7pt} 
\resizebox{\linewidth}{!}{
\begin{tabular}{cccccccc}
\hline
\rowcolor{gray!15} 
\textbf{$\lambda$} & EN & SF & SD & CC & SCD & VIF & RoR \\ \hline
4 & 7.432 & 6.199 & 10.77 & 0.640 & 1.734 & \textbf{0.676} & 2 \\
\rowcolor{gray!5} 
4.8 & 7.410 & 5.931 & 10.54 & 0.637 & \textbf{1.747} & 0.668 & 3 \\
5 & \textbf{7.434} & \textbf{6.235} & \textbf{10.69} & \textbf{0.642} & 1.736 & 0.675 & 1 \\
\rowcolor{gray!5} 
5.2 & 7.355 & 5.920 & 10.33 & 0.634 & 1.732 & 0.634 & 5 \\
6 & 7.383 & 5.623 & 10.52 & 0.640 & 1.720 & 0.627 & 4 \\ \hline
\end{tabular}
}
\label{tab2}
\end{table}

\textbf{Trade-off Parameter $\lambda_2$:} In Eq.(\ref{eq8}), we introduced the regularization term $\mathcal{L}_s$ to preserve local texture details. To identify an appropriate value for $\lambda_2$ that optimizes fusion performance, we performed interpolation experiments following the strategy outlined in \cite{4,35}. The quantitative results, as reported in Table \ref{tab2}, indicate that the model performs better when $\lambda_2 = 5$.

\subsubsection{Ablation Study}
\begin{figure}[t!]
    \centering
    \includegraphics[scale=0.24]{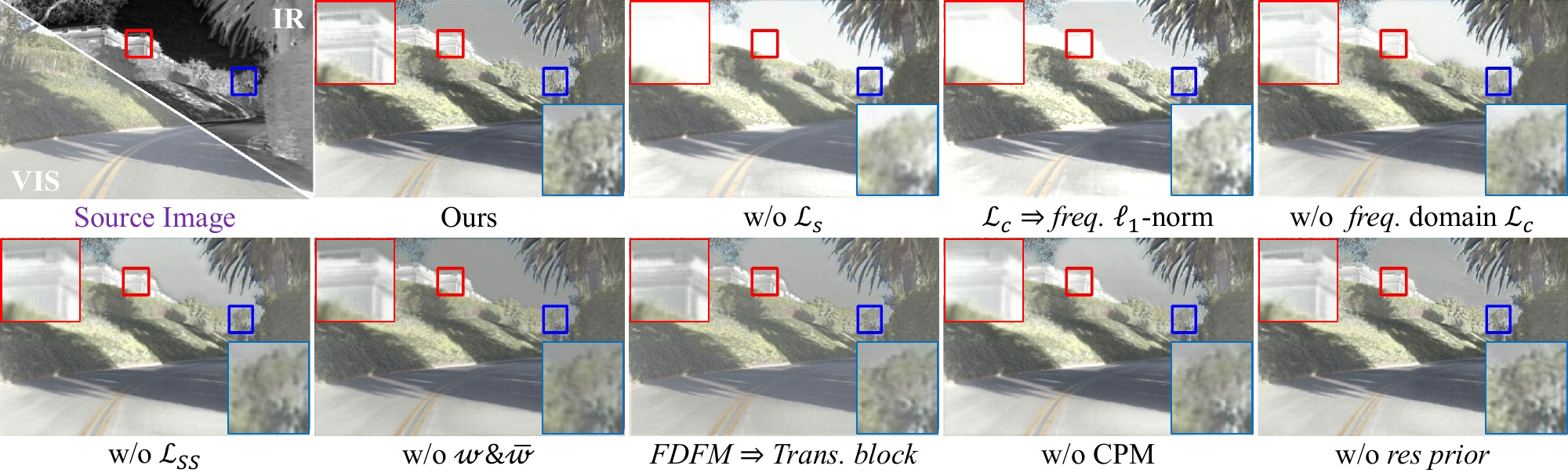}
    \caption{
    A visualization of the ablation experiment. The magnified red boxes show detailed patches of the fusion results.
    }
    \label{fig4}
\end{figure}

We conducted ablation studies across eight cases to evaluate the rationality of each design component on the validation dataset, with quantitative and qualitative results presented in Table \ref{tab3} and Figure \ref{fig4}. In Case 1 (w/o $\mathcal{L}_s$), we removed $\mathcal{L}_s$ to assess its complementary role in supporting the global constraint $\mathcal{L}_c$. In Case 2 ($\mathcal{L}_c \Rightarrow $ \textit{freq.}), we implemented the global constraint solely in the frequency domain, omitting contrastive learning. In Case 3 (w/o \textit{freq.} domain $\mathcal{L}_c$), we constructed contrastive learning in the spatial domain instead. In Case 4 (w/o $w\&\bar{w}$ ), we excluded the adaptive weights $w$, directly enforcing feature constraints between source images. These four cases investigated the rationality of the fusion loss design, revealing that the synergy of frequency-domain global constraints and spatial-domain local constraints preserves texture and contextual information. Notably, frequency-domain contrastive learning driven by adaptive weights $w$ enhances the integration of complementary information. 

\begin{table}[]
\caption{Quantitative results of various ablation studies conducted on the validation dataset. The best values are highlighted in \textbf{bold}.}
\centering
\renewcommand\arraystretch{1.2}
\setlength\tabcolsep{7pt} 
\resizebox{\linewidth}{!}{
\begin{tabular}{ccccccc}
\hline
\rowcolor{gray!15} 
\textbf{Case} & EN & SF & SD & CC & SCD & VIF \\ \hline
w/o $\mathcal{L}_s$ & 7.254 & 4.994 & 10.53 & 0.587 & 1.490 & 0.479 \\
\rowcolor{gray!5} 
$\mathcal{L}_c \Rightarrow $ \textit{freq.} $\ell_1$-norm & 7.356 & 6.040 & \textbf{10.95} & 0.608 & 1.609 & 0.627 \\
w/o \textit{freq.} domain $\mathcal{L}_c$ & 7.371 & 4.906 & 10.36 & 0.605 & 1.576 & 0.462 \\
\rowcolor{gray!5} 
w/o $\mathcal{L}_{SS}$ & 7.336 & 4.697 & 10.24 & 0.634 & 1.650 & 0.547 \\
w/o $w\&\bar{w}$ & 7.175 & 4.105 & 9.96 & \textbf{0.651} & 1.622 & 0.477 \\
\rowcolor{gray!5} 
$FDFM \Rightarrow Trans. block$ & 6.997 & 4.581 & 10.27 & 0.638 & 1.575 & 0.402 \\
w/o CPM & 7.422 & 4.687 & 10.21 & 0.628 & 1.663 & 0.553 \\
\rowcolor{gray!5} 
w/o \textit{res prior} & 7.387 & 5.282 & 10.61 & 0.640 & 1.684 & 0.610 \\
\rowcolor{yellow!10} 
Ours & \textbf{7.434} & \textbf{6.235} & 10.69 & 0.642 & \textbf{1.736} & \textbf{0.675} \\ \hline
\end{tabular}
}
\label{tab3}
\end{table}

\begin{table*}[!t]
\caption{Quantitative results of the proposed method compared with nine other SOTA methods on TNO, MSRS, and M$^3$FD datasets. The best values are highlighted in bold.}
\centering
\renewcommand\arraystretch{1.1}
\setlength\tabcolsep{5pt} 
\resizebox{\linewidth}{!}{
\begin{tabular}{l|ccccc c|ccccc c|ccccc c}
\hline
\rowcolor{gray!15} 
\textbf{Method} & \multicolumn{6}{c|}{\textbf{TNO}} & \multicolumn{6}{c|}{\textbf{MSRS}} & \multicolumn{6}{c}{\textbf{M$^3$FD}} \\ 
\cline{2-19} 
\rowcolor{gray!15} 
 & EN & SF & SD & CC & SCD & VIF & EN & SF & SD & CC & SCD & VIF & EN & SF & SD & CC & SCD & VIF \\ 
\hline
LRRNet \cite{9} & 7.05 & 3.81 & 9.17 & 0.44 & 1.53 & 0.46 & 6.19 & 3.31 & 7.82 & 0.52 & 0.79 & 0.43 & 6.44 & 4.21 & 9.31 & 0.54 & 1.46 & 0.40 \\
\rowcolor{gray!5} 
CS$^2$Fusion \cite{4}& 6.84 & 3.67 & 9.19 & 0.44 & 1.55 & 0.36 & 6.65 & 4.06 & 8.40 & 0.60 & 1.58 & 0.77 & 6.71 & 4.30 & 9.90 & 0.46 & 1.26 & 0.34 \\
Text-IF \cite{7} & 7.21 & 5.18 & 9.54 & 0.45 & 1.69 & 0.70 & 6.74 & 4.67 & 8.52 & 0.60 & \textbf{1.70} & 0.91 & 6.93 & 6.21 & 9.88 & 0.50 & 1.52 & 0.63 \\
\rowcolor{gray!5} 
FISCNet \cite{5} & 6.99 & 4.98 & 9.35 & 0.44 & 1.58 & 0.47 & 6.68 & 4.94 & 8.32 & 0.60 & 1.55 & 0.89 & 6.91 & 6.64 & 10.08 & 0.46 & 1.27 & 0.55 \\
TarDAL \cite{13} & 7.13 & 4.72 & 9.46 & 0.47 & 1.67 & 0.63 & 6.49 & 4.84 & 8.22 & 0.59 & 1.42 & 0.82 & 7.15 & 4.95 & 9.71 & 0.51 & 1.55 & 0.56 \\
\rowcolor{gray!5} 
IGNet \cite{8} & 7.07 & 3.86 & 9.39 & 0.46 & 1.57 & 0.68 & 6.10 & 3.89 & 7.72 & 0.65 & 1.57 & 0.74 & 7.04 & 5.14 & 9.02 & 0.52 & 1.66 & 0.80 \\
SHIP \cite{6} & 6.93 & 4.76 & 9.25 & 0.44 & 1.58 & 0.40 & 6.44 & 4.64 & 8.15 & 0.59 & 1.51 & 0.79 & 6.83 & 6.03 & 10.01 & 0.47 & 1.31 & 0.47 \\
\rowcolor{gray!5} 
Conti-Fuse \cite{10} & 6.87 & 3.99 & 9.28 & 0.45 & 1.59 & 0.38 & 6.67 & 4.50 & 8.38 & 0.60 & 1.64 & 0.81 & 6.76 & 5.54 & 9.98 & 0.48 & 1.37 & 0.43 \\
A$^2$RNet \cite{14} & 7.05 & 3.43 & 9.47 & 0.46 & 1.65 & 0.56 & 6.60 & 3.49 & 8.56 & 0.60 & 1.48 & 0.66 & 6.73 & 3.47 & 9.10 & 0.52 & 1.50 & 0.43 \\
\rowcolor{yellow!10} 
\textbf{Ours} & \textbf{7.44} & \textbf{6.03} & \textbf{10.29} & \textbf{0.52} & \textbf{1.82} & \textbf{1.18} & \textbf{7.28} & \textbf{6.44} & \textbf{10.06} & \textbf{0.61} & 1.51 & \textbf{1.07} & \textbf{7.40} & \textbf{7.14} & \textbf{10.31} & \textbf{0.58} & \textbf{1.85} & \textbf{1.02} \\ 
\hline
\end{tabular}
}
\label{tab4}
\end{table*}

\begin{figure*}[t!]
    \centering
    \includegraphics[scale=0.28]{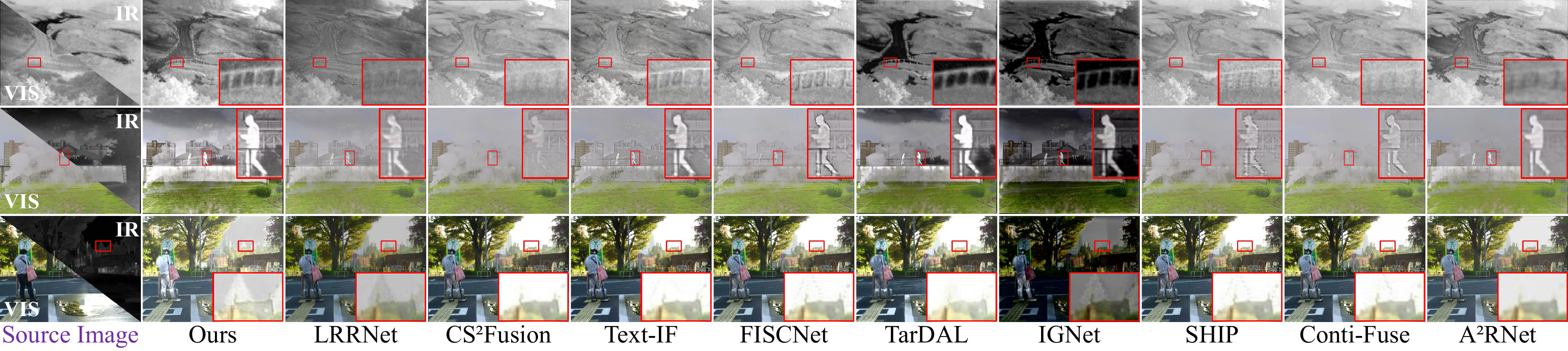}
    \caption{
Visual comparison of different methods on the TNO, MSRS, and M$^3$FD datasets. The proposed method effectively integrates discriminative features to enhance detailed textures and salient objects. The magnified red boxes show detailed patches of the fusion results.
    }
    \label{fig5}
\end{figure*}

In Case 5 (w/o $\mathcal{L}_{SS}$), we removed the auxiliary decoder to evaluate its role in facilitating the capture of differential features. In Case 6 ($FDFM \Rightarrow Trans. block$), we replaced the FDFM with a Transformer block to examine the frequency domain’s capability in capturing global features. In Case 7 (w/o CPM), we excluded the CPM to verify its effectiveness in providing cross-domain cues. In Case 8 (w/o \textit{res prior}), we removed the residual prior branch to assess its contribution to promoting the integration of complementary features. These four cases validated the rationality of the proposed architecture, with results underscoring the effectiveness of frequency-domain feature modeling and the residual prior in enhancing complementary feature integration.

\begin{figure}[t!]
    \centering
    \includegraphics[scale=0.25]{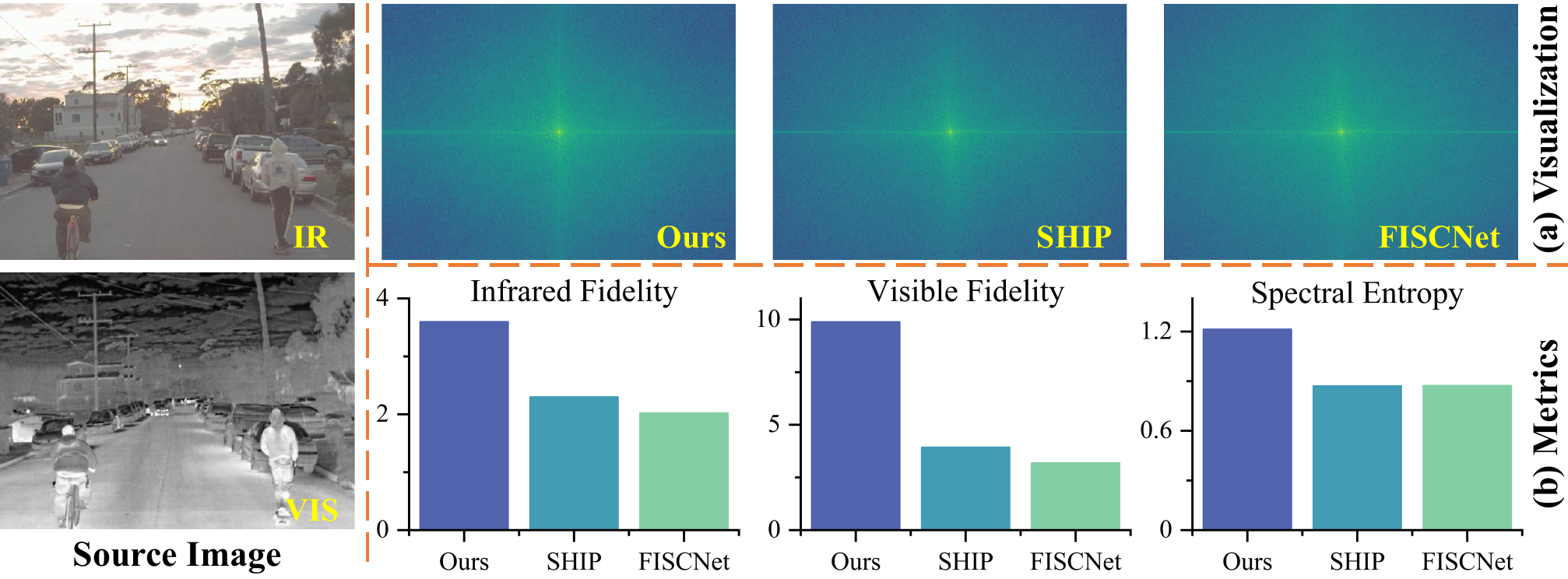}
    \caption{
Demonstration of frequency characteristics of different methods.
    }
    \label{fig8}
\end{figure}

\subsubsection{Fusion result on the test dataset}
Figure \ref{fig5} and Table \ref{tab4} present the qualitative and quantitative results of our proposed method on the test dataset. Our method effectively preserves complementary features from source images through residual prior guidance and global modeling in the frequency domain. Additionally, the adaptive weight-based frequency domain contrastive learning enables our model to better differentiate textural details and high-contrast features. This characteristic allows our method to penetrate through smoke and overexposed environments to better represent scene information—a comprehensive capability that other SOTA methods struggle to achieve.
Notably, while frequency domain modeling approaches like SHIP and FISCNet are effective, they focus primarily on modeling complementary features while neglecting the guidance of residual priors, resulting in a gap in discriminative feature expression compared to our method. Quantitatively, our method's dominant position further confirms its effectiveness. High scores on EN, SD, and SF metrics demonstrate our method's significant capability in integrating textural details, while performance on CC, SCD, and VIF metrics indicates superior integration of complementary information from source images, highlighting its ability to transfer rich information from source images to the fused result. Although our method shows a slight deficiency in SCD scores on MSRS, from an overall perspective, it still demonstrates robustness and generalizability.

\subsection{Extended Experiments}
\subsubsection{Evaluation with Power Spectral Density}
We introduce Power Spectral Density (PSD) as an analytical tool to examine the frequency characteristics of images, facilitating a deeper understanding and evaluation of fusion image quality and performance. Figure \ref{fig8} presents the PSD visualization results (a) and their quantitative analysis (b) for different methods. Compared to frequency-domain-based methods like FISCNet and SHIP, our proposed method exhibits a richer and more uniform energy distribution in the PSD, with notably higher concentration and brightness in the central region. This suggests that our method more effectively preserves thermal target information and texture details. Quantitative results further corroborate this finding: the proposed method dominates in PSD-based metrics, including infrared fidelity and visible fidelity, highlighting its superiority in balancing complementary information from both modalities. Moreover, its advantage in spectral entropy underscores a higher information richness in the frequency distribution. These outcomes demonstrate the effectiveness of our residual-prior-driven and frequency-aware fusion framework in jointly capturing low-frequency global features and high-frequency local details.

\subsubsection{Extension to object detection}
\begin{figure}[t!]
    \centering
    \includegraphics[scale=0.4]{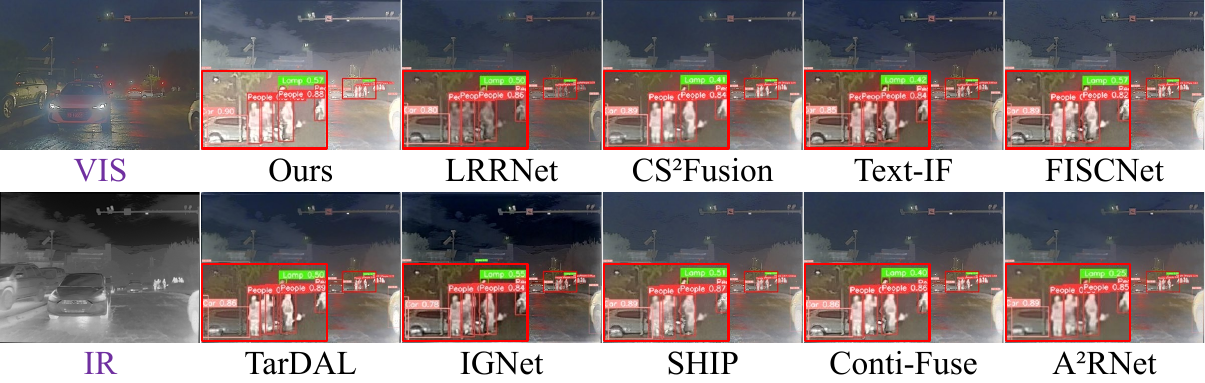}
    \caption{
Visual comparison of detection results across different fused images. The proposed method, by integrating complementary features across modalities, generates high-confidence detection results in complex scenes while delivering visually compelling performance. The magnified red boxes display detailed patches of the detection results.
    }
    \label{fig6}
\end{figure}

Figure \ref{fig6} illustrates the qualitative results of our proposed method on the object detection. Evidently, our method demonstrates superior capability in contextual integration and complementary feature enhancement, effectively facilitating object detection with improved confidence. Furthermore, the per-category detection results presented in Table \ref{tab5} substantiate its efficacy; our method not only achieves the highest mAP@.5 overall, but also exhibits exceptional performance across \textit{'People'}, \textit{'Car'}, \textit{'Bus'}, and \textit{'Motorcycle'} categories. This indicates that the proposed method successfully integrates complex scene information while preserving thermal target saliency without losing textural details, thereby enhancing the detection system's recognition capabilities across diverse object categories.

\begin{table}[!t]
\caption{Quantitative comparison of the proposed method with nine other SOTA methods on the object detection task. The best values are highlighted in bold.}
\centering
\renewcommand\arraystretch{1.3} 
\setlength\tabcolsep{5pt} 
\resizebox{\linewidth}{!}{
\begin{tabular}{lccccccc}
\hline
\rowcolor{gray!15} 
\textbf{Method} & \textit{Peo} & \textit{Car} & \textit{Bus} & \textit{Mot} & \textit{Tru} & \textit{Lam} & mAP@.5 \\
\hline
LRRNet \cite{9}    & 0.784          & 0.909          & 0.889          & 0.685          & 0.831          & 0.8            & 0.816          \\
\rowcolor{gray!5} 
CS$^2$Fusion  \cite{4} & 0.785          & 0.902          & 0.875          & 0.682          & 0.825          & 0.77           & 0.807          \\
Text-IF \cite{7}   & 0.788          & 0.91           & 0.873          & 0.718          & 0.813          & \textbf{0.832} & 0.822          \\
\rowcolor{gray!5} 
FISCNet \cite{5}   & 0.79           & 0.911          & 0.888          & 0.695          & 0.806          & 0.813          & 0.817          \\
TarDAL  \cite{13}   & 0.797          & 0.904          & 0.874          & 0.652          & 0.792          & 0.778          & 0.799          \\
\rowcolor{gray!5} 
IGNet \cite{8}     & 0.767          & 0.855          & 0.739          & 0.452          & 0.662          & 0.565          & 0.673          \\
SHIP  \cite{6}     & 0.79           & 0.909          & 0.876          & 0.686          & 0.808          & 0.811          & 0.813          \\
\rowcolor{gray!5} 
Conti-Fuse \cite{10}& 0.789          & 0.91           & 0.875          & 0.65           & \textbf{0.834} & 0.797          & 0.809          \\
A$^2$RNet \cite{14}    & 0.795          & 0.906          & 0.886          & 0.659          & 0.81           & 0.798          & 0.809          \\
\rowcolor{yellow!10} 
\textbf{Ours} & \textbf{0.796} & \textbf{0.911} & \textbf{0.895} & \textbf{0.725} & 0.806          & 0.8            & \textbf{0.822} \\
\hline
\end{tabular}
}
\label{tab5}
\end{table}

\begin{table}[!t]
\caption{Quantitative comparison between the proposed method and SOTA methods on the Harvard Medical dataset. The best values are highlighted in \textbf{bold}.}
\centering
\renewcommand\arraystretch{1.1} 
\setlength\tabcolsep{10pt} 
\resizebox{\linewidth}{!}{
\begin{tabular}{lcccccc}
\hline
\rowcolor{gray!15} 
\textbf{Method} & EN & SF & SD & CC & SCD & VIF \\
\hline
MATR  \cite{43}    & \textbf{5.15} & 6.86 & 9.41 & 0.75 & 0.28 & 0.34 \\
\rowcolor{gray!5} 
MMIF-INet \cite{44}  & 5.15 & 8.64 & 9.34 & 0.81 & 1.37 & 0.71 \\
MMNet \cite{45}     & 4.74 & 7.46 & \textbf{9.51} & 0.77 & 0.88 & 0.41 \\
\rowcolor{yellow!10} 
\textbf{Ours} & {4.36} & \textbf{9.99} & 9.32 & \textbf{0.81} & \textbf{1.59} & \textbf{0.86} \\
\hline
\end{tabular}
}
\label{tab6}
\end{table}
\begin{figure}[]
    \centering
    \includegraphics[scale=0.23]{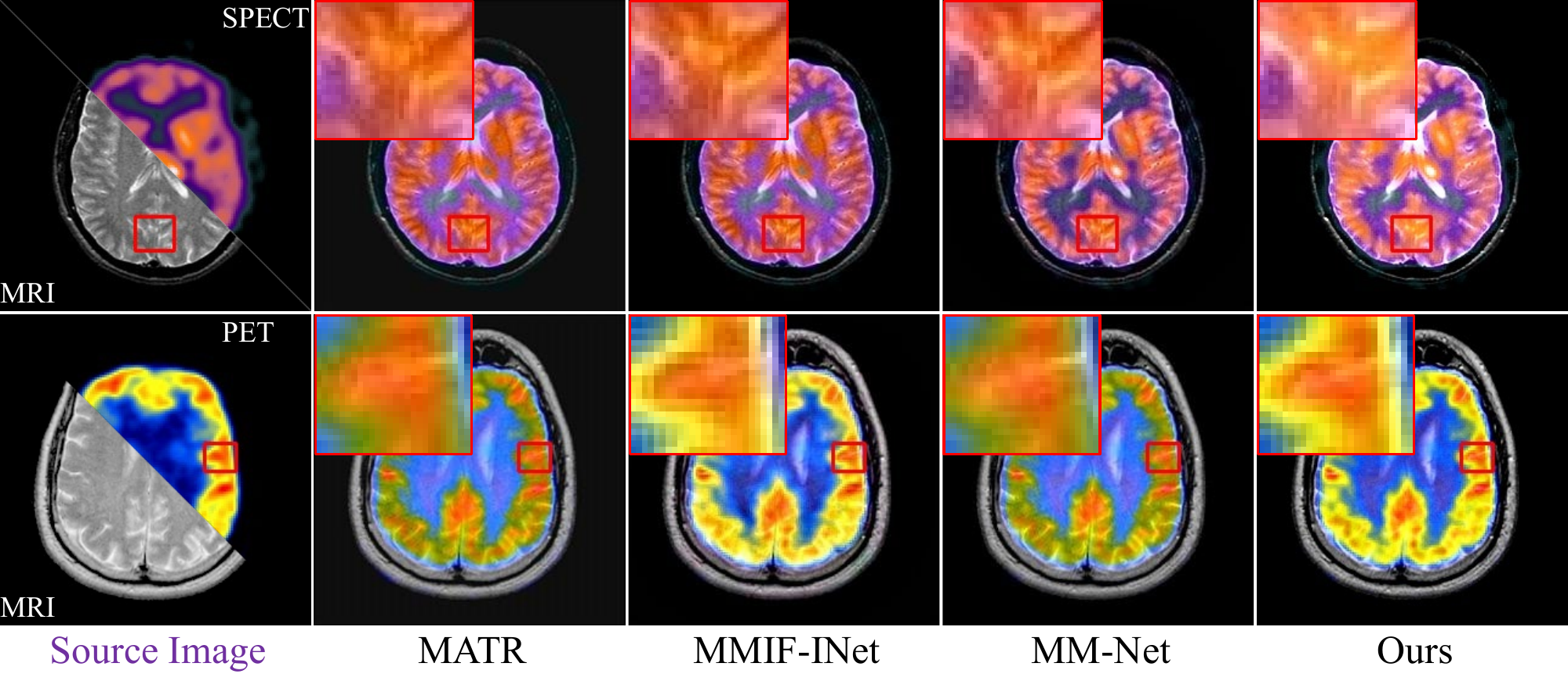}
    \caption{
    Visual comparison of the proposed method with SOTA methods on the Harvard Medical dataset. The magnified red boxes show detailed patches of the fusion results.
    }
    \label{fig7}
\end{figure}

\subsubsection{Extension to MIF}
Furthermore, we extended our proposed method to MIF to validate its effectiveness in multi-modal image fusion tasks. Figure \ref{fig7} and Table \ref{tab6} present both qualitative and quantitative results of our method. Under the guidance of residual priors, our method achieves effective integration of complementary features, preserving detailed textures while highlighting structural information. Consequently, it not only delivers perceptually appealing visual results but also demonstrates statistical advantages in objective metrics. These outcomes verify our method's capability to handle complex scenes, showcasing its adaptability and robustness across various fusion scenarios.

\section{Conclusion}

This research proposes a Residual Prior-driven Frequency-aware Network that integrates global information via frequency-domain convolution and leverages saliency structural priors from residual maps to preserve complementary cross-modal information. The CPM employs a bidirectional closed-loop architecture that retains both local details and salient features. During training, adaptive-weight-based frequency-domain contrastive loss preserves complementary features globally, while saliency structure loss uses an auxiliary decoder to capture modality-specific differences. These design elements enable the method to highlight salient objects while preserving fine-grained details, with extensive experiments validating its effectiveness on both static statistics and high-level vision tasks.

\begin{acks}
This work was supported by National Natural Science Foundation of China under Grants No.62162065 and 61761045; Yunnan Fundamental Research Projects No.202201AT070167; Joint Special Project Research Foundation of Yunnan Province No.202401BF070001-023; Yunnan Province Visual and Cultural Innovation Team No.202505-AS350009. 
\end{acks}


\bibliographystyle{ACM-Reference-Format}
\bibliography{./samples/Cite.bib}


\pagestyle{empty}
\end{document}